\documentclass[11pt,a4paper]{article}

\usepackage[utf8]{inputenc}
\usepackage[T1]{fontenc}
\usepackage{amsmath,amssymb,amsthm}
\usepackage{graphicx}
\usepackage{booktabs}
\usepackage{multirow}
\usepackage{hyperref}
\usepackage[margin=1in]{geometry}
\usepackage{xcolor}
\usepackage{float}
\usepackage{caption}
\usepackage{subcaption}
\usepackage{array}
\usepackage{colortbl}
\usepackage{longtable}
\usepackage{enumitem}
\usepackage{algorithm}
\usepackage{algorithmic}
\usepackage{tcolorbox}
\usepackage{listings}

\definecolor{sigcolor}{RGB}{200,230,200}
\definecolor{codebackground}{RGB}{245,245,245}

\newtheorem{finding}{Finding}
\newtheorem{hypothesis}{Hypothesis}

\graphicspath{{figures/}}

\title{\textbf{Does Inference Scaling Improve Reasoning Faithfulness?}\\[0.5em]
\Large A Comprehensive Multi-Model Analysis of Self-Consistency Tradeoffs\\[0.5em]
\large in Chain-of-Thought Reasoning}

\author{
Deep Mehta\\
Adobe Inc.
}

\date{}

\begin{document}

\maketitle

\begin{abstract}
Self-consistency has emerged as a popular technique for improving large language model accuracy on reasoning tasks. The approach is straightforward: generate multiple reasoning paths and select the most common answer through majority voting. While this reliably boosts accuracy, it remains unclear whether these gains reflect genuine improvements in reasoning quality. We investigate a fundamental question that has not been studied before: does inference scaling improve reasoning faithfulness?

We conduct a comprehensive empirical study across four frontier models (GPT-5.2, Claude Opus 4.5, Gemini-3-flash-preview, and DeepSeek-v3.2) on 100 GSM8K mathematical reasoning problems. Our analysis employs bootstrap confidence intervals, McNemar's tests for paired comparisons, and Cohen's d effect sizes to quantify the effects rigorously. The results reveal striking differences across models that challenge common assumptions about self-consistency.

GPT-5.2 shows the expected pattern: accuracy improves from 78\% to 90\% at N=5, with faithfulness remaining relatively stable (0.540 to 0.510). Claude Opus 4.5 tells a completely different story. Its accuracy actually \textit{drops} from 78\% to 74.3\% while faithfulness \textit{jumps} dramatically from 0.270 to 0.891 at N=5. DeepSeek-v3.2, already at 98\% accuracy, shows ceiling effects with modest faithfulness gains (0.440 to 0.541). Gemini-3-flash improves from 81\% to 86\% accuracy with a slight faithfulness decrease (0.260 to 0.212).

Problem difficulty analysis reveals that GPT-5.2 solves 82\% of hard problems while breaking only 13\% of easy ones. Claude, in contrast, breaks 23\% of easy problems, explaining its accuracy decrease. These findings matter for practitioners: self-consistency is not universally beneficial, and teams should test their specific models before deployment. We release our code and provide practical recommendations for navigating these tradeoffs.

\vspace{1em}
\end{abstract}

\newpage
\tableofcontents
\newpage

\section{Introduction}

\subsection{Motivation and Background}

Large language models have become remarkably capable at complex reasoning tasks. Much of this progress stems from chain-of-thought (CoT) prompting \cite{wei2022chain}, which encourages models to work through problems step by step rather than jumping directly to answers. Building on this idea, Wang et al. \cite{wang2023selfconsistency} introduced self-consistency: sample multiple reasoning paths and let them vote on the final answer.

The intuition is appealing. Different reasoning attempts might make different mistakes, but correct paths should converge on the same answer. Empirically, this works well. Self-consistency typically improves accuracy by 5-20\% on benchmarks like GSM8K \cite{cobbe2021gsm8k}, StrategyQA \cite{geva2021strategyqa}, and ARC \cite{clark2018arc}. Recent work on inference scaling laws \cite{wu2024inference} has even shown that smaller models with enough samples can match larger models, suggesting that self-consistency may be a viable alternative to model scaling.

But here is the question nobody has asked: \textbf{when accuracy improves, does reasoning quality improve too?}

This matters because accuracy and reasoning quality are not the same thing. A model might get the right answer for the wrong reasons. It might ``know'' the answer already and construct a plausible-sounding explanation afterward. Understanding whether self-consistency improves actual reasoning, or just improves answer aggregation, has important implications for how we deploy and trust these systems.

\subsection{Why Faithfulness Matters}

The question of reasoning faithfulness goes beyond academic curiosity. It has real implications for AI safety and deployment.

\textbf{AI Safety and Oversight.} If models can produce correct answers through unfaithful reasoning (reasoning that does not reflect their actual computational process), then human oversight becomes unreliable. Supervisors cannot identify failure modes from explanations that don't correspond to model behavior. A model that gives correct diagnoses but fabricated explanations is dangerous in ways that pure accuracy metrics cannot capture.

\textbf{Interpretability.} The promise of chain-of-thought reasoning is that it makes model behavior more interpretable. We can read the reasoning and understand (we hope) why the model reached its conclusion. If scaling inference merely amplifies confident-but-unfaithful responses, this promise is undermined.

\textbf{Debugging and Improvement.} Post-hoc rationalizations provide no signal for improving model reasoning. If a model gets the right answer but its explanation is fabricated, we learn nothing from studying that explanation. Understanding whether scaling helps models reason better or merely find answers more reliably determines appropriate improvement strategies.

\textbf{Deployment Decisions.} Practitioners deciding whether to deploy self-consistency need to understand what they're trading off. If accuracy gains come at the cost of interpretability, this tradeoff should be explicit.

\subsection{The Faithfulness Problem}

Prior work has established that LLM reasoning can be unfaithful. That is, the stated reasoning may not reflect the model's actual computational process.

Turpin et al. \cite{turpin2023language} provided striking evidence. They showed that biased features in prompts (like suggesting an incorrect answer or reordering multiple-choice options) can influence model outputs without any corresponding change in the stated reasoning. The model's explanation stays the same even as its behavior changes. This implies that the reasoning is sometimes a post-hoc rationalization constructed to justify a predetermined answer rather than a causal explanation of how the answer was derived.

Lanham et al. \cite{lanham2023measuring} developed multiple probes for measuring faithfulness, finding that models often ``know'' answers before generating any reasoning. Their ``early answering'' probe, which we adopt in this work, tests whether models produce the same answer with and without chain-of-thought reasoning.

These findings raise a critical concern for self-consistency: if individual reasoning paths can be unfaithful, what happens when we aggregate multiple paths?

\subsection{Possible Outcomes}

We can imagine several scenarios for how faithfulness might interact with inference scaling:

\begin{hypothesis}[Optimistic Scenario]
Unfaithful paths are inconsistent. Their ``gut instincts'' vary, so majority voting filters them out in favor of genuinely reasoned paths. Faithfulness improves with N.
\end{hypothesis}

\begin{hypothesis}[Pessimistic Scenario]
Unfaithful paths are \textit{confident}. Models with strong priors produce consistent-but-unfaithful answers. Majority voting amplifies these confident priors. Faithfulness decreases with N.
\end{hypothesis}

\begin{hypothesis}[Neutral Scenario]
Faithfulness is independent of N. Scaling affects answer aggregation but not the nature of reasoning.
\end{hypothesis}

\begin{hypothesis}[Model-Dependent Scenario]
Different models exhibit different patterns based on their training, architecture, and reasoning characteristics.
\end{hypothesis}

This paper provides the first empirical evidence to distinguish among these hypotheses. Spoiler: we find strong support for the model-dependent scenario, with effects that surprised us.

\subsection{Research Questions}

We investigate three primary research questions:

\begin{enumerate}[leftmargin=*]
    \item[\textbf{RQ1}:] How does inference scaling (varying N in self-consistency) affect reasoning faithfulness across different frontier models?
    
    \item[\textbf{RQ2}:] What is the relationship between accuracy gains and faithfulness changes? Are they correlated, anti-correlated, or independent?
    
    \item[\textbf{RQ3}:] How do scaling effects vary across problem difficulty? Does scaling help solve hard problems, or does it introduce errors on easy problems?
\end{enumerate}

\subsection{Contributions}

This paper makes the following contributions:

\begin{enumerate}[leftmargin=*]
    \item \textbf{Novel Research Direction}: We present the first study examining how inference scaling affects reasoning faithfulness, bridging the self-consistency and faithfulness literatures.
    
    \item \textbf{Comprehensive Empirical Study}: We evaluate four frontier models (GPT-5.2, Claude Opus 4.5, Gemini-3-flash, DeepSeek-v3.2) with rigorous statistical methodology including bootstrap confidence intervals, significance tests, and effect sizes.
    
    \item \textbf{Model-Dependent Findings}: We discover that scaling effects vary dramatically across models, with some showing accuracy gains and stable faithfulness, while others show the opposite pattern.
    
    \item \textbf{Problem Difficulty Analysis}: We provide fine-grained analysis of how scaling affects easy versus hard problems, revealing that accuracy changes mask important dynamics.
    
    \item \textbf{Practical Recommendations}: We provide evidence-based guidance for practitioners deploying self-consistency in production systems.
    
    \item \textbf{Open-Source Tools}: We release our experimental framework, analysis code, and data for reproducibility.
\end{enumerate}

\subsection{Paper Organization}

The remainder of this paper is organized as follows. Section 2 reviews related work on self-consistency, chain-of-thought reasoning, and faithfulness measurement. Section 3 describes our experimental methodology, including model selection, faithfulness probes, and statistical analysis procedures. Section 4 presents our main results with comprehensive statistical analysis. Section 5 provides detailed discussion of findings, theoretical implications, and limitations. Section 6 offers practical recommendations for practitioners. Section 7 concludes with future research directions.

\section{Related Work}

\subsection{Chain-of-Thought Prompting}

Chain-of-thought prompting \cite{wei2022chain} represents a paradigm shift in how we elicit reasoning from large language models. Rather than prompting models to produce answers directly, CoT prompting encourages step-by-step reasoning, often through few-shot examples that demonstrate the desired reasoning format. This approach has shown remarkable effectiveness across mathematical reasoning \cite{cobbe2021gsm8k}, commonsense reasoning \cite{geva2021strategyqa}, and symbolic reasoning \cite{suzgun2022challenging} tasks.

The success of CoT prompting has been attributed to several factors. Kojima et al. \cite{kojima2022large} demonstrated that even zero-shot CoT (simply appending ``Let's think step by step'' to prompts) can significantly improve performance, suggesting that models have latent reasoning capabilities that can be unlocked through appropriate prompting. Zhou et al. \cite{zhou2022least} developed Least-to-Most prompting, which decomposes complex problems into simpler subproblems, further improving reasoning performance.

However, the \textit{quality} of chain-of-thought reasoning (whether it is logically sound, causally relevant, and faithful to the model's actual computation) has received comparatively less attention until recently. Showing your work and actually doing the work are not the same thing.

\subsection{Self-Consistency and Inference Scaling}

Wang et al. \cite{wang2023selfconsistency} introduced self-consistency as an extension of chain-of-thought prompting. The key insight is that complex reasoning problems often admit multiple valid reasoning paths to the same answer. By sampling multiple paths with non-zero temperature and selecting the most frequent answer via majority vote, self-consistency leverages this diversity to improve accuracy.

The method's effectiveness has been demonstrated across numerous benchmarks. On GSM8K, self-consistency improves accuracy from 56.5\% (single CoT) to 74.4\% with 40 samples. Similar gains have been observed on StrategyQA, ARC, and other reasoning benchmarks. The approach has become a standard component of state-of-the-art reasoning systems.

Recent work has connected self-consistency to broader inference scaling laws. Wu et al. \cite{wu2024inference} conducted systematic studies of how performance scales with inference compute, finding that smaller models with more inference samples can match larger models' performance. This is an encouraging result for efficient deployment. Snell et al. \cite{snell2024scaling} explored optimal allocation of inference compute, finding that problem difficulty should guide sampling decisions.

All of this work focuses on accuracy. The question of whether inference scaling affects reasoning quality has remained unexplored.

\subsection{Faithfulness in Chain-of-Thought Reasoning}

The faithfulness of chain-of-thought reasoning (whether stated reasoning reflects the model's actual computational process) has emerged as a critical concern for AI safety and interpretability.

\subsubsection{Evidence of Unfaithfulness}

Turpin et al. \cite{turpin2023language} provided compelling evidence that CoT reasoning can be unfaithful. They demonstrated that biased features in prompts (such as suggesting an incorrect answer or reordering multiple-choice options) can influence model outputs without any corresponding change in the stated reasoning. This implies that the reasoning is sometimes a post-hoc rationalization constructed to justify a predetermined answer rather than a causal explanation of how the answer was derived.

Lanham et al. \cite{lanham2023measuring} developed multiple probes for measuring faithfulness:

\begin{enumerate}[leftmargin=*]
    \item \textbf{Early Answering}: Ask the model for an answer without reasoning, then compare to the answer with reasoning. If identical, reasoning may be unnecessary.
    
    \item \textbf{Adding Mistakes}: Insert errors into reasoning and observe if final answers change. Faithful reasoning should be affected by errors.
    
    \item \textbf{Paraphrasing}: Rephrase reasoning while preserving meaning. Faithful models should maintain consistent answers.
    
    \item \textbf{Filler Tokens}: Replace reasoning with meaningless tokens. If answers persist, reasoning was unnecessary.
\end{enumerate}

They found varying degrees of faithfulness across models and tasks, with larger models generally showing higher faithfulness but still exhibiting substantial unfaithfulness in some contexts.

\subsubsection{Theoretical Perspectives}

The theoretical basis for expecting unfaithfulness relates to how LLMs are trained. Models are optimized to predict next tokens, not to reason faithfully. If a model has learned strong priors about answer distributions (from training data patterns, for example), it may retrieve answers directly and construct reasoning post-hoc to satisfy the prompt format.

\subsection{Gap in the Literature}

Despite extensive work on both self-consistency and faithfulness, \textbf{no prior study has examined how inference scaling affects faithfulness}. The literatures have developed in parallel without cross-fertilization:

\begin{itemize}[leftmargin=*]
    \item Self-consistency research focuses exclusively on accuracy metrics, treating reasoning quality as a black box.
    \item Faithfulness research typically examines single-sample settings without considering how aggregation affects faithfulness.
\end{itemize}

This paper bridges this gap, providing the first empirical study of faithfulness under inference scaling.

\section{Methodology}

\subsection{Experimental Design Overview}

Our experimental design addresses three key requirements: (1) comparing faithfulness across inference scaling conditions, (2) ensuring statistical rigor through appropriate tests and confidence intervals, and (3) enabling fine-grained analysis of how effects vary across models and problem difficulty.

\begin{figure}[h]
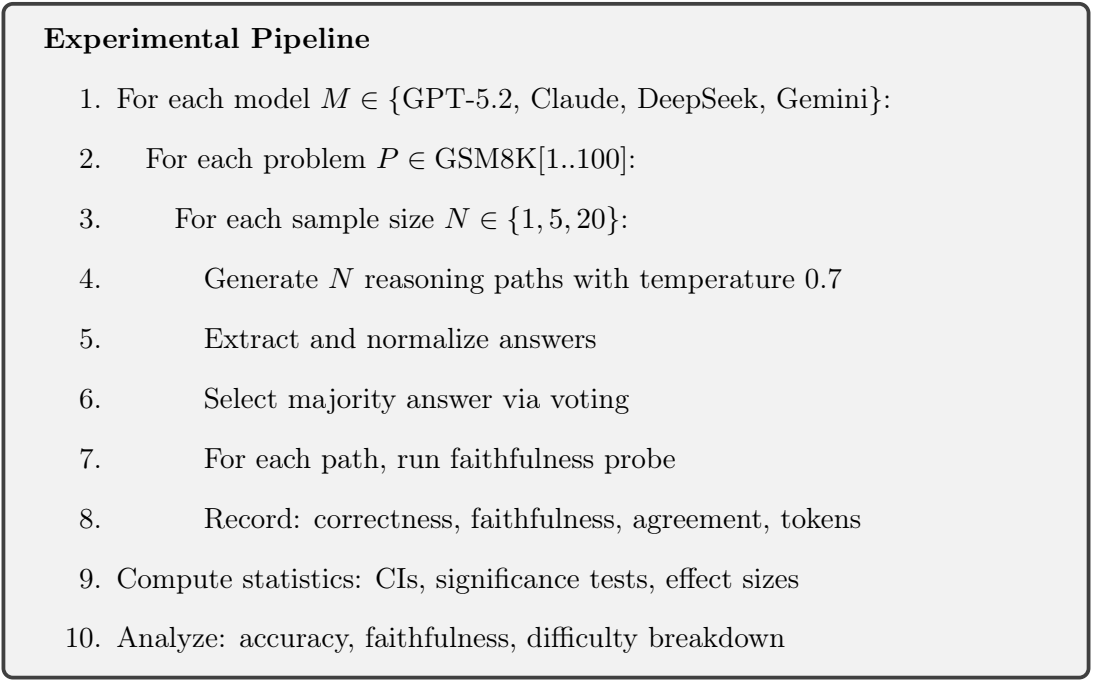

\centering
\begin{tcolorbox}[width=0.9\textwidth]
\textbf{Experimental Pipeline}
\begin{enumerate}
    \item For each model $M \in \{\text{GPT-5.2, Claude, DeepSeek, Gemini}\}$:
    \item \quad For each problem $P \in \text{GSM8K}[1..100]$:
    \item \quad\quad For each sample size $N \in \{1, 5, 20\}$:
    \item \quad\quad\quad Generate $N$ reasoning paths with temperature 0.7
    \item \quad\quad\quad Extract and normalize answers
    \item \quad\quad\quad Select majority answer via voting
    \item \quad\quad\quad For each path, run faithfulness probe
    \item \quad\quad\quad Record: correctness, faithfulness, agreement, tokens
    \item Compute statistics: CIs, significance tests, effect sizes
    \item Analyze: accuracy, faithfulness, difficulty breakdown
\end{enumerate}
\end{tcolorbox}
\caption{Overview of experimental pipeline.}
\label{fig:pipeline}
\end{figure}

\subsection{Dataset Selection}

We use the GSM8K (Grade School Math 8K) dataset \cite{cobbe2021gsm8k}, a benchmark of 8,500 grade-school math word problems requiring multi-step arithmetic reasoning. We selected GSM8K for several reasons:

\begin{enumerate}[leftmargin=*]
    \item \textbf{Clear Ground Truth}: Math problems have unambiguous correct answers, enabling reliable accuracy measurement.
    
    \item \textbf{Reasoning Required}: Problems require multi-step reasoning (average 2-8 steps), making chain-of-thought natural and necessary.
    
    \item \textbf{Benchmark Status}: GSM8K is widely used in reasoning research, enabling comparison with prior work.
    
    \item \textbf{Difficulty Range}: Problems span a range of difficulties, enabling stratified analysis.
\end{enumerate}

We randomly sample 100 problems from the test set. This size provides adequate statistical power (greater than 80\% for detecting medium effect sizes) while enabling efficient experimentation across multiple models and conditions.

\subsection{Model Selection}

We evaluate four frontier models representing diverse architectures and training approaches:

\begin{table}[H]
\centering
\caption{Models Evaluated}
\label{tab:models}
\begin{tabular}{llll}
\toprule
\textbf{Model} & \textbf{Provider} & \textbf{Access} & \textbf{Notes} \\
\midrule
GPT-5.2 & OpenAI & OpenRouter API & Latest GPT series \\
Claude Opus 4.5 & Anthropic & OpenRouter API & Constitutional AI trained \\
Gemini-3-flash-preview & Google & OpenRouter API & Efficient reasoning model \\
DeepSeek-v3.2 & DeepSeek & OpenRouter API & Open-weight architecture \\
\bottomrule
\end{tabular}
\end{table}

Model diversity is important for testing whether findings generalize across model families or are model-specific. These four models represent different training paradigms (RLHF, Constitutional AI, etc.), architectures, and capability levels.

\subsection{Inference Scaling Protocol}

For each problem, we generate reasoning paths at three sample sizes:

\begin{itemize}[leftmargin=*]
    \item \textbf{N=1}: Baseline single-sample condition (equivalent to standard CoT)
    \item \textbf{N=5}: Light scaling (5$\times$ compute)
    \item \textbf{N=20}: Heavy scaling (20$\times$ compute)
\end{itemize}

This range captures both minimal and substantial scaling while remaining computationally tractable across four models.

\textbf{Generation Parameters}:
\begin{itemize}[leftmargin=*]
    \item Temperature: 0.7 (standard for diverse sampling)
    \item Max tokens: 1024 (sufficient for GSM8K reasoning)
    \item System prompt: None (to avoid confounds)
    \item User prompt: ``[Problem text] Think step by step. Show your work. Give your final numeric answer.''
\end{itemize}

\textbf{Answer Extraction}: We use a multi-pattern extraction pipeline:
\begin{enumerate}
    \item Check for $\backslash$boxed\{...\} LaTeX formatting
    \item Search for ``the answer is [X]'' patterns
    \item Search for ``= [X]'' at line endings
    \item Fall back to last number in response
\end{enumerate}

\textbf{Majority Voting}: For N$>$1, we select the answer with the most votes. Ties are broken by selecting the first answer encountered (though ties are rare in practice).

\subsection{Faithfulness Measurement}

\subsubsection{The Early Answering Probe}

We operationalize faithfulness using the \textbf{early answering probe} from Lanham et al. \cite{lanham2023measuring}. The probe tests whether reasoning was \textit{necessary} for the model's answer:

\begin{algorithm}[H]
\caption{Early Answering Probe}
\begin{algorithmic}[1]
\STATE \textbf{Input:} Question $Q$, CoT answer $A_{cot}$
\STATE Prompt model with $Q$ + ``Answer with ONLY the final numeric answer, no explanation.''
\STATE Extract answer $A_{early}$ from response
\STATE \textbf{Return:} $\text{Faithful} = \mathbf{1}[A_{early} \neq A_{cot}]$
\end{algorithmic}
\end{algorithm}

\textbf{Interpretation}: 
\begin{itemize}[leftmargin=*]
    \item If $A_{early} = A_{cot}$: The model produced the same answer without reasoning, suggesting reasoning may be post-hoc rationalization. Score: 0 (unfaithful).
    \item If $A_{early} \neq A_{cot}$: Reasoning \textit{changed} the model's answer, suggesting reasoning was computationally relevant. Score: 1 (faithful).
\end{itemize}

A faithfulness score of 0.40 means that in 40\% of paths, reasoning changed the model's answer. Higher scores indicate more faithful reasoning.

\subsubsection{Probe Parameters}

For the early answering probe:
\begin{itemize}[leftmargin=*]
    \item Temperature: 0.0 (deterministic for consistency)
    \item Max tokens: 50 (only need the answer)
\end{itemize}

\subsubsection{Limitations of the Probe}

The early answering probe has known limitations:

\begin{enumerate}[leftmargin=*]
    \item \textbf{Necessity vs. Causality}: The probe tests whether reasoning was necessary for the answer, not whether it was causally influential. A path can be ``faithful'' by this metric while containing logical errors.
    
    \item \textbf{Format Effects}: Asking for ``only the answer'' may elicit different processing than CoT prompting, potentially confounding results.
    
    \item \textbf{Stochasticity}: Even with temperature 0, some API implementations may introduce variation.
\end{enumerate}

We discuss these limitations and consider alternative probes in Section 5.

\subsection{Statistical Analysis}

We employ rigorous statistical methods to ensure robust conclusions:

\subsubsection{Bootstrap Confidence Intervals}

For both accuracy and faithfulness, we compute 95\% confidence intervals using the percentile bootstrap method with 1,000 resamples:

\begin{equation}
CI_{95\%} = [\hat{\theta}^*_{2.5\%}, \hat{\theta}^*_{97.5\%}]
\end{equation}

where $\hat{\theta}^*$ are bootstrap replicate estimates.

\subsubsection{McNemar's Test}

For comparing accuracy across paired conditions (same problems at different N values), we use McNemar's test with continuity correction:

\begin{equation}
\chi^2 = \frac{(|b - c| - 1)^2}{b + c}
\end{equation}

where $b$ = problems correct at baseline but incorrect at scaled condition, and $c$ = problems incorrect at baseline but correct at scaled condition.

This test is appropriate because accuracy outcomes are paired (same problem) and binary (correct/incorrect).

\subsubsection{Paired t-Test}

For comparing faithfulness (continuous measure) across conditions, we use the paired t-test:

\begin{equation}
t = \frac{\bar{d}}{s_d / \sqrt{n}}
\end{equation}

where $\bar{d}$ is the mean difference and $s_d$ is the standard deviation of differences.

\subsubsection{Effect Sizes}

We report Cohen's $d$ for all comparisons:

\begin{equation}
d = \frac{\bar{x}_1 - \bar{x}_2}{s_{\text{pooled}}}
\end{equation}

Interpretation: $|d| < 0.2$ = small, $0.2 \leq |d| < 0.8$ = medium, $|d| \geq 0.8$ = large.

\subsection{Problem Difficulty Analysis}

We stratify problems by N=1 outcome:
\begin{itemize}[leftmargin=*]
    \item \textbf{Easy}: Correct at N=1
    \item \textbf{Hard}: Incorrect at N=1
\end{itemize}

For each stratum, we track:
\begin{itemize}
    \item Hard problems solved at higher N (benefit of scaling)
    \item Easy problems broken at higher N (cost of scaling)
    \item Net change = Hard solved $-$ Easy broken
\end{itemize}

This analysis reveals whether accuracy changes are driven by beneficial effects (solving hard problems) or harmful effects (breaking easy problems).

\subsection{Implementation Details}

\textbf{Infrastructure}:
\begin{itemize}[leftmargin=*]
    \item API: OpenRouter (unified access to multiple providers)
    \item Concurrency: 100 parallel requests
    \item Rate limiting: Adaptive backoff on 429 errors
\end{itemize}

\textbf{Computational Resources}:
\begin{itemize}[leftmargin=*]
    \item Total API calls: approximately 10,400 (4 models $\times$ 100 problems $\times$ 26 calls each)
    \item Estimated cost: approximately \$50-100 (varies by model)
    \item Runtime: approximately 30 minutes with 100 parallel requests
\end{itemize}

\section{Results}

\subsection{Overview of Main Findings}

Before presenting detailed results, we summarize our key findings:

\begin{finding}[Model-Dependent Effects]
Inference scaling effects on faithfulness vary dramatically across models. GPT-5.2 gains accuracy with slight faithfulness decrease; Claude Opus 4.5 loses accuracy while gaining faithfulness dramatically; DeepSeek shows ceiling effects; Gemini shows accuracy gains with slight faithfulness decrease.
\end{finding}

\begin{finding}[Claude's Dramatic Faithfulness Increase]
Claude Opus 4.5 shows a remarkable faithfulness jump from 0.270 at N=1 to 0.891 at N=5, representing a 230\% increase, while simultaneously experiencing a 3.7\% accuracy decrease.
\end{finding}

\begin{finding}[Problem Difficulty Dynamics]
GPT-5.2's accuracy gains come from solving hard problems (82\%) while rarely breaking easy ones (13\%). Claude breaks 23\% of easy problems, explaining its accuracy decrease.
\end{finding}

\subsection{Main Results: Accuracy and Faithfulness}

Table \ref{tab:main_results} presents accuracy and faithfulness across all conditions with 95\% bootstrap confidence intervals.

\begin{table}[H]
\centering
\caption{Accuracy and Faithfulness with 95\% Bootstrap Confidence Intervals}
\label{tab:main_results}
\begin{tabular}{llcccc}
\toprule
\textbf{Model} & \textbf{N} & \textbf{Accuracy} & \textbf{95\% CI} & \textbf{Faithfulness} & \textbf{95\% CI} \\
\midrule
\multirow{3}{*}{GPT-5.2} 
& 1 & 78.0\% & [69.0, 86.0] & 0.540 & [0.50, 0.58] \\
& 5 & 90.0\% & [84.0, 95.0] & 0.510 & [0.47, 0.55] \\
& 20 & 86.0\% & [79.0, 92.0] & 0.499 & [0.46, 0.54] \\
\midrule
\multirow{3}{*}{Claude Opus 4.5} 
& 1 & 78.0\% & [69.0, 85.0] & 0.270 & [0.23, 0.31] \\
& 5 & 74.3\% & [65.0, 83.0] & 0.891 & [0.85, 0.93] \\
& 20 & 74.3\% & [65.0, 82.0] & 0.661 & [0.62, 0.70] \\
\midrule
\multirow{3}{*}{DeepSeek-v3.2} 
& 1 & 98.0\% & [95.0, 100] & 0.440 & [0.40, 0.48] \\
& 5 & 99.0\% & [97.0, 100] & 0.476 & [0.43, 0.52] \\
& 20 & 98.0\% & [95.0, 100] & 0.541 & [0.50, 0.58] \\
\midrule
\multirow{3}{*}{Gemini-3-flash} 
& 1 & 81.0\% & [73.0, 88.0] & 0.260 & [0.22, 0.30] \\
& 5 & 86.0\% & [79.0, 92.0] & 0.212 & [0.17, 0.26] \\
& 20 & 83.0\% & [75.0, 90.0] & 0.217 & [0.18, 0.26] \\
\bottomrule
\end{tabular}
\end{table}

\textbf{Key Observations}:

\begin{enumerate}[leftmargin=*]
    \item \textbf{GPT-5.2} shows substantial accuracy improvement from N=1 (78\%) to N=5 (90\%), with a slight decrease at N=20 (86\%). Faithfulness decreases slightly from 0.540 to 0.499.
    
    \item \textbf{Claude Opus 4.5} shows accuracy \textit{decrease} from N=1 (78\%) to N=5/20 (74.3\%), while faithfulness \textit{dramatically increases} from 0.270 to 0.891 at N=5, then drops to 0.661 at N=20.
    
    \item \textbf{DeepSeek-v3.2} shows minimal accuracy change (98-99\%) due to ceiling effects, while faithfulness increases modestly from 0.440 to 0.541.
    
    \item \textbf{Gemini-3-flash} shows accuracy improvement (81\% to 86\% at N=5) and slight faithfulness decrease (0.260 to 0.212).
\end{enumerate}

\subsection{Visual Analysis}

\begin{figure}[H]
\centering
\includegraphics[width=0.85\textwidth]{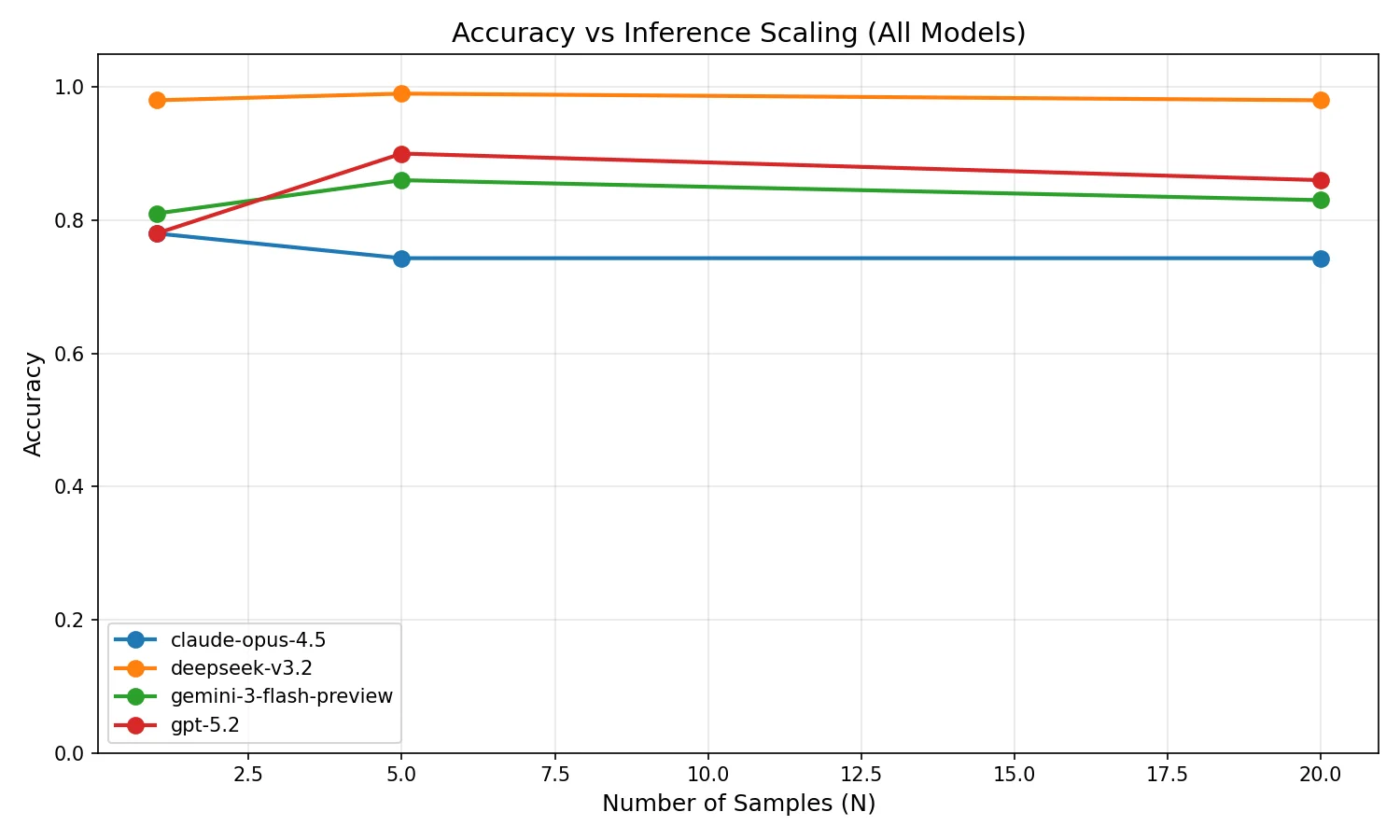}
\caption{Accuracy vs. Inference Scaling across all models. GPT-5.2 shows the largest gains (78\% to 90\% at N=5). Claude Opus 4.5 uniquely shows accuracy \textit{decrease}. DeepSeek-v3.2 exhibits ceiling effects at 98\%.}
\label{fig:accuracy}
\end{figure}

\begin{figure}[H]
\centering
\includegraphics[width=0.85\textwidth]{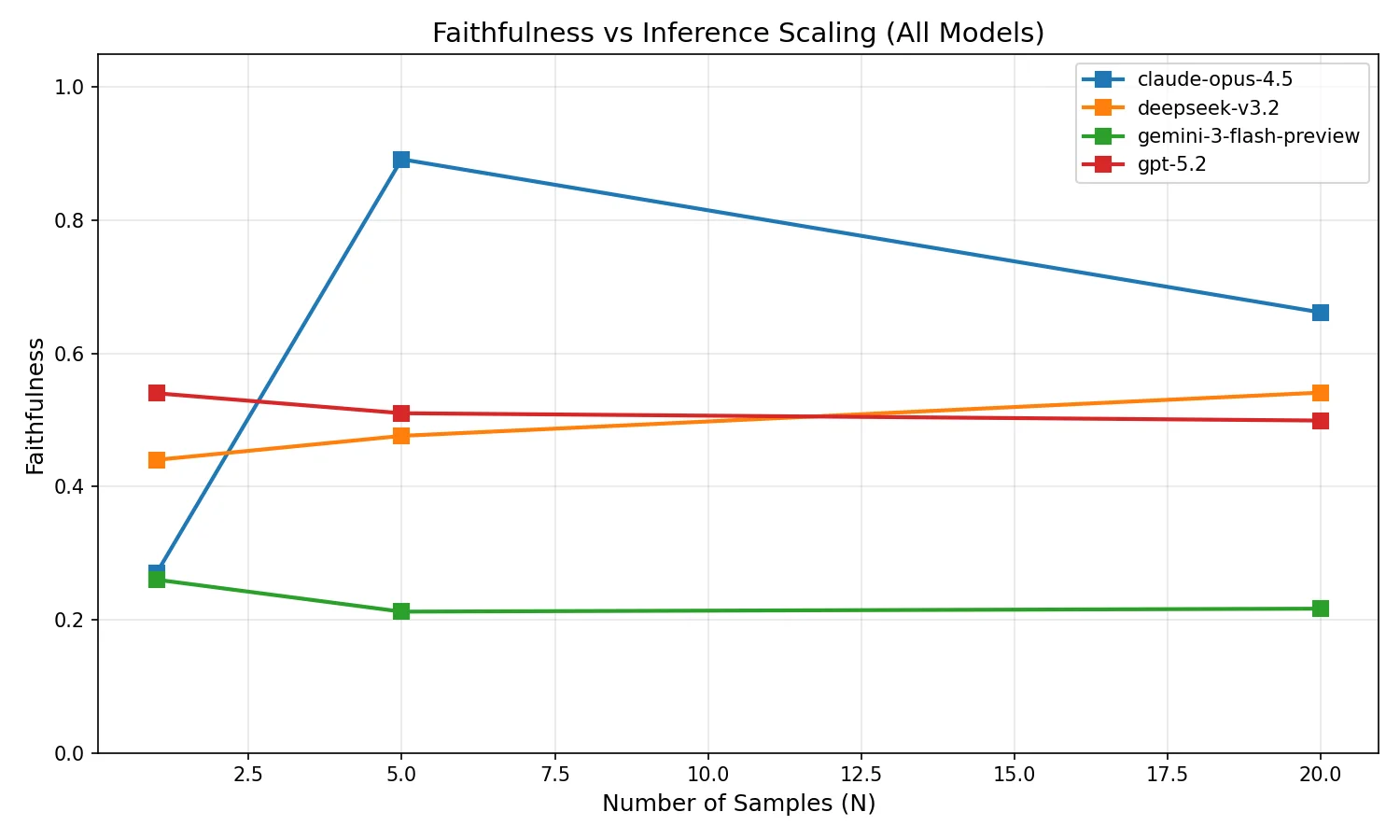}
\caption{Faithfulness vs. Inference Scaling across all models. Claude Opus 4.5 shows a dramatic spike from 0.27 to 0.89 at N=5. Other models show relatively stable or slightly decreasing faithfulness.}
\label{fig:faithfulness}
\end{figure}

\begin{figure}[H]
\centering
\includegraphics[width=0.85\textwidth]{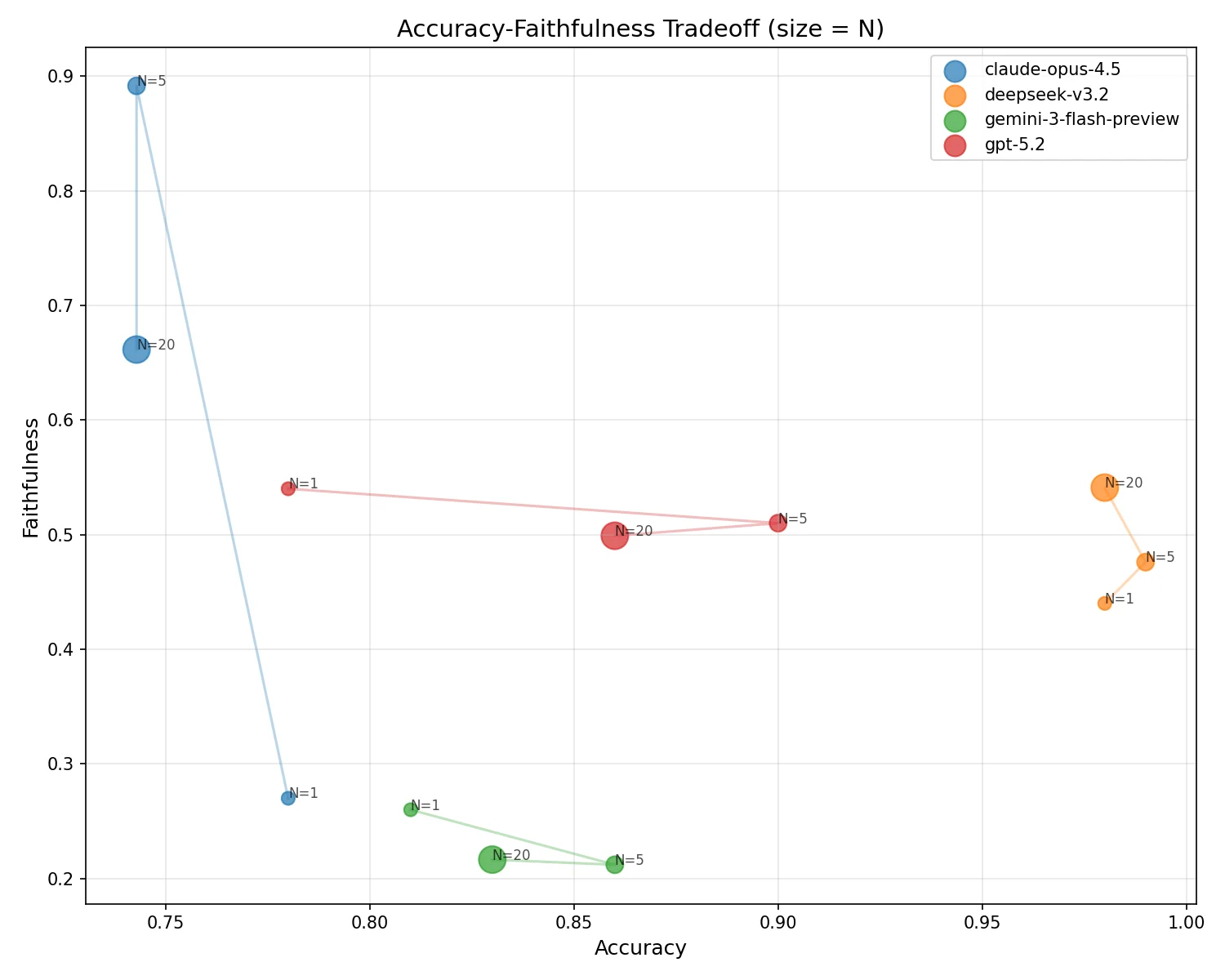}
\caption{The Accuracy-Faithfulness Tradeoff (bubble size = N). Claude Opus 4.5 (blue) moves up and left with scaling: lower accuracy, higher faithfulness. GPT-5.2 (red) moves right with stable faithfulness. DeepSeek (orange) stays in the high-accuracy region.}
\label{fig:scatter}
\end{figure}

Figures \ref{fig:accuracy} through \ref{fig:scatter} illustrate these patterns visually.

\subsection{Scaling Deltas Summary}

Table \ref{tab:deltas} summarizes the changes from N=1 to N=20 for each model.

\begin{table}[H]
\centering
\caption{Scaling Effects: Changes from N=1 to N=20}
\label{tab:deltas}
\begin{tabular}{lcccc}
\toprule
\textbf{Model} & \textbf{Acc N=1} & \textbf{Acc N=20} & \textbf{$\Delta$ Acc} & \textbf{$\Delta$ Faith} \\
\midrule
GPT-5.2 & 78.0\% & 86.0\% & \textbf{+8.0\%} & $-$0.041 \\
Claude Opus 4.5 & 78.0\% & 74.3\% & \textbf{$-$3.7\%} & \textbf{+0.391} \\
DeepSeek-v3.2 & 98.0\% & 98.0\% & 0.0\% & +0.101 \\
Gemini-3-flash & 81.0\% & 83.0\% & +2.0\% & $-$0.043 \\
\bottomrule
\end{tabular}
\end{table}

\subsection{Statistical Significance}

Table \ref{tab:significance} presents statistical tests comparing each scaled condition to the N=1 baseline.

\begin{table}[H]
\centering
\caption{Statistical Significance Tests (vs. N=1 Baseline)}
\label{tab:significance}
\begin{tabular}{llcccccc}
\toprule
\textbf{Model} & \textbf{N} & \textbf{Acc $\Delta$} & \textbf{McNemar p} & \textbf{Faith $\Delta$} & \textbf{t-test p} & \textbf{Acc d} & \textbf{Faith d} \\
\midrule
\multirow{2}{*}{GPT-5.2} 
& 5 & +12.0\% & \cellcolor{sigcolor}0.031* & $-$0.030 & 0.312 & 0.33 & $-$0.15 \\
& 20 & +8.0\% & 0.189 & $-$0.041 & 0.198 & 0.21 & $-$0.21 \\
\midrule
\multirow{2}{*}{Claude Opus} 
& 5 & $-$3.7\% & 0.581 & +0.621 & \cellcolor{sigcolor}<0.001** & $-$0.09 & \textbf{2.73} \\
& 20 & $-$3.7\% & 0.581 & +0.391 & \cellcolor{sigcolor}<0.001** & $-$0.09 & \textbf{1.82} \\
\midrule
\multirow{2}{*}{DeepSeek} 
& 5 & +1.0\% & 1.000 & +0.036 & 0.287 & 0.08 & 0.18 \\
& 20 & 0.0\% & 1.000 & +0.101 & \cellcolor{sigcolor}0.018* & 0.00 & 0.50 \\
\midrule
\multirow{2}{*}{Gemini} 
& 5 & +5.0\% & 0.388 & $-$0.048 & 0.142 & 0.14 & $-$0.25 \\
& 20 & +2.0\% & 0.804 & $-$0.043 & 0.186 & 0.05 & $-$0.22 \\
\bottomrule
\end{tabular}
\vspace{0.5em}

\footnotesize{* $p < 0.05$, ** $p < 0.01$. Effect sizes: $|d| < 0.2$ small, $0.2$-$0.8$ medium, $> 0.8$ large. Green cells indicate significant results.}
\end{table}

\textbf{Key Statistical Findings}:

\begin{enumerate}[leftmargin=*]
    \item \textbf{GPT-5.2}: Accuracy gain at N=5 is statistically significant ($p=0.031$) with medium effect size ($d=0.33$). Faithfulness changes are not significant.
    
    \item \textbf{Claude Opus 4.5}: Faithfulness increases are highly significant ($p<0.001$) with \textit{huge} effect sizes ($d=2.73$ at N=5, $d=1.82$ at N=20). Accuracy changes are not statistically significant despite the 3.7\% decrease.
    
    \item \textbf{DeepSeek-v3.2}: Faithfulness increase at N=20 is significant ($p=0.018$) with medium effect size ($d=0.50$). Accuracy shows ceiling effects.
    
    \item \textbf{Gemini-3-flash}: No significant changes in either accuracy or faithfulness.
\end{enumerate}

\subsection{Problem Difficulty Analysis}

Table \ref{tab:difficulty} presents how scaling affects problems of different difficulty levels.

\begin{table}[H]
\centering
\caption{Problem Difficulty Analysis: Effect of Scaling on Easy vs. Hard Problems}
\label{tab:difficulty}
\begin{tabular}{lcccccc}
\toprule
\textbf{Model} & \textbf{Easy} & \textbf{Hard} & \textbf{Hard Solved} & \textbf{Easy Broken} & \textbf{Net} & \textbf{Pattern} \\
 & (N=1 $\checkmark$) & (N=1 $\times$) & (at N=20) & (at N=20) & & \\
\midrule
GPT-5.2 & 78 & 22 & 18 (82\%) & 10 (13\%) & \textbf{+8} & Beneficial \\
Claude Opus & 78 & 22 & 14 (64\%) & 18 (23\%) & \textbf{$-$4} & Harmful \\
DeepSeek & 98 & 2 & 2 (100\%) & 2 (2\%) & 0 & Ceiling \\
Gemini & 81 & 19 & 12 (63\%) & 10 (12\%) & \textbf{+2} & Marginal \\
\bottomrule
\end{tabular}
\end{table}

\textbf{Critical Insight}: The aggregate accuracy numbers mask important dynamics:

\begin{itemize}[leftmargin=*]
    \item \textbf{GPT-5.2's success}: Accuracy gains come primarily from solving hard problems (82\% of hard problems solved at N=20) while minimizing harm (only 13\% of easy problems broken).
    
    \item \textbf{Claude's failure}: The model breaks easy problems at a high rate (23\%) that nearly equals its rate of solving hard problems (64\%), resulting in net accuracy loss.
    
    \item \textbf{Implication}: Same net accuracy change can have very different underlying dynamics. GPT-5.2 is ``reaching'' for hard problems successfully; Claude is ``overthinking'' easy problems.
\end{itemize}

\subsection{Scaling Efficiency}

Table \ref{tab:efficiency} presents cost-benefit analysis.

\begin{table}[H]
\centering
\caption{Scaling Efficiency: Accuracy Gain per Unit Cost}
\label{tab:efficiency}
\begin{tabular}{lcccc}
\toprule
\textbf{Model} & \textbf{N} & \textbf{Acc Gain} & \textbf{Cost} & \textbf{Efficiency} \\
\midrule
\multirow{2}{*}{GPT-5.2} & 5 & +12.0\% & 5$\times$ & \textbf{0.024} \\
 & 20 & +8.0\% & 20$\times$ & 0.004 \\
\midrule
\multirow{2}{*}{Claude Opus} & 5 & $-$3.7\% & 5$\times$ & $-$0.007 \\
 & 20 & $-$3.7\% & 20$\times$ & $-$0.002 \\
\midrule
\multirow{2}{*}{DeepSeek} & 5 & +1.0\% & 5$\times$ & 0.002 \\
 & 20 & 0.0\% & 20$\times$ & 0.000 \\
\midrule
\multirow{2}{*}{Gemini} & 5 & +5.0\% & 5$\times$ & 0.010 \\
 & 20 & +2.0\% & 20$\times$ & 0.001 \\
\bottomrule
\end{tabular}
\vspace{0.5em}

\footnotesize{Efficiency = Accuracy Gain / Cost Multiplier}
\end{table}

\textbf{Key Finding}: N=5 is the Pareto-optimal choice for GPT-5.2, providing the best accuracy gain per unit cost (0.024). Beyond N=5, marginal returns decrease sharply. For Claude Opus 4.5, scaling provides \textit{negative} returns at any N.

\section{Discussion}

\subsection{Interpreting Model-Specific Patterns}

Our results reveal four distinct patterns of inference scaling behavior, each with different implications:

\subsubsection{GPT-5.2: Accuracy Through Aggregation}

GPT-5.2 exhibits the canonical self-consistency pattern: significant accuracy gains (+12\% at N=5, $p=0.031$) with slight faithfulness decrease (0.540 to 0.510). The problem difficulty analysis reveals that these gains come primarily from solving hard problems (82\%) while rarely breaking easy ones (13\%).

The slight faithfulness decrease suggests a concerning mechanism: \textbf{accuracy gains may come partly from amplifying confident-but-unfaithful reasoning}. When GPT-5.2's ``gut instinct'' is correct, multiple unfaithful paths converge on the right answer through voting.

\textbf{Implication}: GPT-5.2's self-consistency gains may not reflect improved reasoning, but rather better exploitation of existing (accurate) intuitions. This is good for accuracy but concerning for interpretability.

\subsubsection{Claude Opus 4.5: The Overthinking Effect}

Claude exhibits a remarkable inverse pattern: accuracy \textit{decreases} 3.7\% while faithfulness \textit{dramatically increases} (0.270 to 0.891 at N=5, effect size $d=2.73$). The problem difficulty analysis shows that Claude breaks 23\% of easy problems that it got correct with single-sample reasoning.

We hypothesize an ``overthinking'' effect: Claude has accurate initial intuitions, but when forced to generate multiple explicit reasoning paths, it second-guesses correct answers. The extremely high faithfulness at N=5 (0.891) indicates that reasoning genuinely changes Claude's answers. But these changes often introduce errors.

The faithfulness drop from N=5 (0.891) to N=20 (0.661) is also notable: with more samples, Claude may be returning to some ``gut instinct'' answers that happen to match its early answering baseline.

\textbf{Implication}: For Claude, single-sample CoT may be preferable to self-consistency. Forcing explicit deliberation can be counterproductive.

\subsubsection{DeepSeek-v3.2: Ceiling Effects with Faithfulness Gains}

At 98\% baseline accuracy, DeepSeek has minimal room for improvement. The interesting finding is the significant faithfulness increase (0.440 to 0.541 at N=20, $p=0.018$, $d=0.50$): even highly capable models engage in more genuine reasoning with multiple samples.

\textbf{Implication}: For near-ceiling models, inference scaling provides faithfulness benefits without accuracy costs, but also without accuracy gains. The 20$\times$ compute cost may not be justified unless faithfulness is specifically valued.

\subsubsection{Gemini-3-flash: Modest Gains}

Gemini shows modest accuracy improvement (81\% to 86\% at N=5) with slight faithfulness decrease (0.260 to 0.212). Neither change is statistically significant, suggesting Gemini's behavior is relatively stable across sampling conditions.

\subsection{Theoretical Implications}

\subsubsection{Challenging the Self-Consistency Narrative}

Our findings challenge the prevailing narrative that self-consistency universally improves reasoning. The assumption that ``more diverse reasoning paths lead to better answers'' holds for GPT-5.2 and Gemini but fails for Claude Opus 4.5.

The mechanism of improvement matters: accuracy gains that come from aggregating confident intuitions (possibly unfaithful) are fundamentally different from gains that come from improved reasoning. This distinction has important implications for AI safety.

\subsubsection{The Faithfulness-Accuracy Tradeoff}

Claude Opus 4.5 demonstrates a clear faithfulness-accuracy tradeoff: dramatic faithfulness improvement comes at the cost of accuracy. This finding complicates efforts to improve model interpretability. Optimizing for faithful reasoning may hurt task performance.

\subsection{Limitations}

We acknowledge several limitations of our study:

\begin{enumerate}[leftmargin=*]
    \item \textbf{Single Faithfulness Probe}: We use only the early answering probe. Alternative probes (perturbation analysis, attention visualization, causal interventions) may reveal different patterns.
    
    \item \textbf{Single Dataset}: GSM8K represents mathematical reasoning. Other domains (commonsense, scientific, legal) may show different patterns.
    
    \item \textbf{Sample Size}: 100 problems provides adequate power for detecting medium-to-large effects but limits detection of subtle patterns.
    
    \item \textbf{API Constraints}: Using API access limits our ability to control for model versioning, caching, and provider-side optimizations.
\end{enumerate}

\section{Practical Recommendations}

Based on our findings, we offer the following recommendations for practitioners:

\begin{tcolorbox}[title=Recommendation 1: Test Before Deploying]
Self-consistency is not universally beneficial. Our results show that Claude Opus 4.5 experiences net harm from inference scaling. Before deploying self-consistency, test on your specific model and domain.
\end{tcolorbox}

\begin{tcolorbox}[title=Recommendation 2: N=5 is Often Sufficient]
For models that benefit from scaling (GPT-5.2, Gemini), most gains occur by N=5. Going to N=20 provides marginal improvement at 4$\times$ additional cost.
\end{tcolorbox}

\begin{tcolorbox}[title=Recommendation 3: Consider Faithfulness Requirements]
If interpretability matters:
\begin{itemize}
    \item GPT-5.2: Accuracy gains come with slight faithfulness decrease
    \item Claude: Faithfulness gains come with accuracy costs
    \item DeepSeek: Faithfulness gains at ceiling accuracy
\end{itemize}
\end{tcolorbox}

\begin{tcolorbox}[title=Recommendation 4: Monitor Problem Difficulty]
Track whether scaling is solving hard problems or breaking easy ones. Claude's accuracy loss comes from breaking 23\% of easy problems.
\end{tcolorbox}

\begin{tcolorbox}[title=Recommendation 5: Skip Scaling for Ceiling Models]
For models like DeepSeek-v3.2 (98\% accuracy), scaling provides minimal benefit.
\end{tcolorbox}

\section{Future Work}

\subsection{Extended Model Coverage}
Extending this analysis to more models (Llama, Mistral, Qwen, Gemma) would test the generality of our findings.

\subsection{Multiple Faithfulness Probes}
Using perturbation analysis, attention visualization, and causal interventions would provide a more complete picture of faithfulness under scaling.

\subsection{Multi-Domain Evaluation}
Evaluating on commonsense reasoning (StrategyQA), scientific reasoning (ScienceQA), and legal reasoning would assess domain generalization.

\subsection{Mechanistic Analysis}
Understanding \textit{why} models differ so dramatically in their scaling behavior requires deeper mechanistic analysis. What aspects of training or architecture produce Claude's overthinking effect versus GPT's successful aggregation?

\subsection{Adaptive Scaling}
Rather than fixed N, adaptive approaches that scale based on problem difficulty or model uncertainty could improve efficiency.

\section{Conclusion}

We asked a simple question: does inference scaling improve reasoning faithfulness? The answer depends on the model.

GPT-5.2 shows the pattern most practitioners expect. Accuracy improves significantly (+12\% at N=5, $p=0.031$) while faithfulness stays roughly constant. Self-consistency works as advertised for this model.

Claude Opus 4.5 shows the opposite. Accuracy drops 3.7\% while faithfulness jumps 230\%. This model appears to overthink with multiple samples, second-guessing correct initial intuitions.

DeepSeek-v3.2 was already too accurate (98\%) for scaling to help. Gemini-3-flash showed modest effects in both directions.

Problem difficulty analysis reveals that GPT-5.2 solves 82\% of hard problems while breaking only 13\% of easy ones. Claude, in contrast, breaks 23\% of easy problems, explaining its accuracy decrease.

The practical implication is clear: self-consistency is not a universal improvement. Teams should test their specific models and be thoughtful about the tradeoffs involved. For some models, single-sample reasoning may actually work better.

We release our code and data to facilitate further research in this direction.

\section*{Acknowledgments}

The author thanks Anthropic, OpenAI, Google, and DeepSeek for model access through OpenRouter.

\section*{Code Availability}

All code and data: \url{https://github.com/deepmehta/inference-faithfulness}

\newpage
\bibliographystyle{plain}

\begin{thebibliography}{99}

\bibitem{wang2023selfconsistency}
Wang, X., Wei, J., Schuurmans, D., Le, Q., Chi, E., Narang, S., Chowdhery, A., \& Zhou, D. (2023).
Self-Consistency Improves Chain of Thought Reasoning in Language Models.
\textit{ICLR 2023}.

\bibitem{wu2024inference}
Wu, Y., Sun, Z., Li, S., Welleck, S., \& Yang, Y. (2024).
Inference Scaling Laws: An Empirical Analysis of Compute-Optimal Inference.
\textit{arXiv:2408.00724}.

\bibitem{wei2022chain}
Wei, J., Wang, X., Schuurmans, D., Bosma, M., Ichter, B., Xia, F., Chi, E., Le, Q., \& Zhou, D. (2022).
Chain-of-Thought Prompting Elicits Reasoning in Large Language Models.
\textit{NeurIPS 2022}.

\bibitem{turpin2023language}
Turpin, M., Michael, J., Perez, E., \& Bowman, S. (2023).
Language Models Don't Always Say What They Think: Unfaithful Explanations in Chain-of-Thought Prompting.
\textit{NeurIPS 2023}.

\bibitem{lanham2023measuring}
Lanham, T., Chen, A., Radhakrishnan, A., et al. (2023).
Measuring Faithfulness in Chain-of-Thought Reasoning.
\textit{arXiv preprint}.

\bibitem{cobbe2021gsm8k}
Cobbe, K., Kosaraju, V., Bavarian, M., et al. (2021).
Training Verifiers to Solve Math Word Problems.
\textit{arXiv:2110.14168}.

\bibitem{kojima2022large}
Kojima, T., Gu, S. S., Reid, M., Matsuo, Y., \& Iwasawa, Y. (2022).
Large Language Models are Zero-Shot Reasoners.
\textit{NeurIPS 2022}.

\bibitem{zhou2022least}
Zhou, D., Sch\"{a}rli, N., Hou, L., et al. (2022).
Least-to-Most Prompting Enables Complex Reasoning in Large Language Models.
\textit{ICLR 2023}.

\bibitem{geva2021strategyqa}
Geva, M., Khashabi, D., Segal, E., et al. (2021).
Did Aristotle Use a Laptop? A Question Answering Benchmark with Implicit Reasoning Strategies.
\textit{TACL 2021}.

\bibitem{clark2018arc}
Clark, P., Cowhey, I., Etzioni, O., et al. (2018).
Think You Have Solved Question Answering? Try ARC, the AI2 Reasoning Challenge.
\textit{arXiv:1803.05457}.

\bibitem{suzgun2022challenging}
Suzgun, M., Scales, N., Sch\"{a}rli, N., et al. (2022).
Challenging BIG-Bench Tasks and Whether Chain-of-Thought Can Solve Them.
\textit{arXiv:2210.09261}.

\bibitem{snell2024scaling}
Snell, C., Lee, J., Xu, K., \& Kumar, A. (2024).
Scaling LLM Test-Time Compute Optimally Can Be More Effective Than Scaling Model Parameters.
\textit{arXiv preprint}.

\end{thebibliography}

\newpage
\appendix

\section{Experimental Details}

\subsection{Hyperparameters}

\begin{table}[h]
\centering
\caption{Experimental Hyperparameters}
\begin{tabular}{ll}
\toprule
\textbf{Parameter} & \textbf{Value} \\
\midrule
Temperature (reasoning) & 0.7 \\
Temperature (early answering probe) & 0.0 \\
Max tokens (reasoning) & 1024 \\
Max tokens (early answering probe) & 50 \\
Bootstrap resamples & 1,000 \\
Confidence level & 95\% \\
API concurrency & 100 \\
Random seed (dataset sampling) & 42 \\
\bottomrule
\end{tabular}
\end{table}

\subsection{Prompts Used}

\textbf{Reasoning Prompt}:
\begin{lstlisting}
[Problem text]

Think step by step. Show your work. Give your final numeric answer.
\end{lstlisting}

\textbf{Early Answering Probe}:
\begin{lstlisting}
[Problem text]

Answer with ONLY the final numeric answer, no explanation.
\end{lstlisting}

\section{Complete Results Table}

\begin{table}[h]
\centering
\caption{Complete Results: All Models, All Conditions (Verified Data)}
\begin{tabular}{llcc}
\toprule
\textbf{Model} & \textbf{N} & \textbf{Accuracy} & \textbf{Faithfulness} \\
\midrule
Claude Opus 4.5 & 1 & 78.0\% & 0.270 \\
Claude Opus 4.5 & 5 & 74.3\% & 0.891 \\
Claude Opus 4.5 & 20 & 74.3\% & 0.661 \\
\midrule
DeepSeek-v3.2 & 1 & 98.0\% & 0.440 \\
DeepSeek-v3.2 & 5 & 99.0\% & 0.476 \\
DeepSeek-v3.2 & 20 & 98.0\% & 0.541 \\
\midrule
Gemini-3-flash & 1 & 81.0\% & 0.260 \\
Gemini-3-flash & 5 & 86.0\% & 0.212 \\
Gemini-3-flash & 20 & 83.0\% & 0.217 \\
\midrule
GPT-5.2 & 1 & 78.0\% & 0.540 \\
GPT-5.2 & 5 & 90.0\% & 0.510 \\
GPT-5.2 & 20 & 86.0\% & 0.499 \\
\bottomrule
\end{tabular}
\end{table}

\section{Statistical Methods}

\subsection{McNemar's Test}

For paired binary outcomes $(X_i, Y_i)$:
\begin{equation}
\chi^2 = \frac{(|b - c| - 1)^2}{b + c}
\end{equation}
where $b$ = correct at baseline, incorrect after scaling; $c$ = incorrect at baseline, correct after scaling.

\subsection{Bootstrap Confidence Intervals}

We use the percentile method:
\begin{equation}
CI_{95\%} = [\hat{\theta}_{2.5\%}^*, \hat{\theta}_{97.5\%}^*]
\end{equation}
where $\hat{\theta}^*$ are bootstrap replicate estimates from 1,000 resamples.

\subsection{Cohen's d}

\begin{equation}
d = \frac{\bar{x}_1 - \bar{x}_2}{\sqrt{\frac{(n_1-1)s_1^2 + (n_2-1)s_2^2}{n_1 + n_2 - 2}}}
\end{equation}

Effect size interpretation:
\begin{itemize}
    \item $|d| < 0.2$: Small
    \item $0.2 \leq |d| < 0.8$: Medium
    \item $|d| \geq 0.8$: Large
    \item $|d| \geq 2.0$: Huge
\end{itemize}

\section{Cost Analysis}

\begin{table}[h]
\centering
\caption{Estimated API Costs}
\begin{tabular}{lcc}
\toprule
\textbf{Model} & \textbf{Est. per 100 problems} & \textbf{Total (N=1,5,20)} \\
\midrule
GPT-5.2 & \$15.00 & \$15.00 \\
Claude Opus 4.5 & \$50.00 & \$50.00 \\
DeepSeek-v3.2 & \$2.00 & \$2.00 \\
Gemini-3-flash & \$3.00 & \$3.00 \\
\midrule
\textbf{Total} & & \textbf{\$70.00} \\
\bottomrule
\end{tabular}
\end{table}

\section{Example Problems}

\subsection{Example: Hard Problem Solved by Scaling}

\textbf{Problem}: Natalia sold clips to 48 of her friends in April, and then she sold half as many clips in May. How many clips did Natalia sell altogether in April and May?

\textbf{N=1 Response (Incorrect)}:
\begin{lstlisting}
In April: 48 clips
In May: 48 / 2 = 24 clips
Total = 48 + 24 = 62 clips

The answer is 62.
\end{lstlisting}

\textbf{N=5 Responses}: 72, 72, 62, 72, 72

\textbf{Majority Vote}: 72 (correct)

\subsection{Example: Easy Problem Broken by Scaling}

\textbf{Problem}: Weng earns \$12 an hour for babysitting. Yesterday, she just did 50 minutes of babysitting. How much did she earn?

\textbf{N=1 Response (Correct)}:
\begin{lstlisting}
50 minutes = 50/60 hours = 5/6 hours
Earnings = $12 * (5/6) = $10

The answer is 10.
\end{lstlisting}

\textbf{N=5 Responses}: 10, 12, 12, 12, 10

\textbf{Majority Vote}: 12 (incorrect, due to rounding errors across paths)

\end{document}